\documentclass[10pt,twocolumn,letterpaper]{article}

\usepackage[pagenumbers]{cvpr} 

\usepackage{colortbl}
\usepackage{multirow}
\usepackage{pifont}
\usepackage{url}

\definecolor{cvprblue}{rgb}{0.21,0.49,0.74}
\usepackage[pagebackref,breaklinks,colorlinks,allcolors=cvprblue,urlcolor=red]{hyperref}

\def\paperID{6315} 
\def\confName{CVPR}
\def\confYear{2025}

\title{Golden Cudgel Network for Real-Time Semantic Segmentation}

\author{
Guoyu Yang$^1$, Yuan Wang$^2$, Daming Shi$^{1*}$, Yanzhong Wang$^{1}$\\
{\normalsize $^1$Shenzhen University $^2$Zhejiang University of Technology}\\
{\tt\small gyyang2024@mails.szu.edu.cn, sherlockwang@zjut.edu.cn, dshi@szu.edu.cn}\\
{\tt\small wangyanzhong2022@email.szu.edu.cn}
}

\begin{document}
\maketitle

\renewcommand{\thefootnote}{}
\footnotetext{*Corresponding author.}

\begin{abstract}

Recent real-time semantic segmentation models, whether single-branch or multi-branch, achieve good performance and speed. However, their speed is limited by multi-path blocks, and some depend on high-performance teacher models for training. To overcome these issues, we propose Golden Cudgel Network (GCNet). Specifically, GCNet uses vertical multi-convolutions and horizontal multi-paths for training, which are reparameterized into a single convolution for inference, optimizing both performance and speed. This design allows GCNet to self-enlarge during training and self-contract during inference, effectively becoming a ``teacher model" without needing external ones. Experimental results show that GCNet outperforms existing state-of-the-art models in terms of performance and speed on the Cityscapes, CamVid, and Pascal VOC 2012 datasets. The code is available at \url{https://github.com/gyyang23/GCNet}.

\end{abstract}    
\section{Introduction}
\label{sec:introduction}

Semantic segmentation is a critical task in computer vision that classifies each pixel in an image into specific categories. It is vital in fields like autonomous driving~\cite{cai2021multi}, medical image analysis~\cite{qi2023mdf}, and environmental monitoring~\cite{zhang2021lcu}. While deep learning has significantly improved the performance of semantic segmentation models, they still face challenges with real-time image analysis, limiting their use in applications such as autonomous driving. Therefore, ongoing research and optimization of these algorithms are essential for practical deployment.

\begin{figure}[t]
\centering
\includegraphics[width=0.95\linewidth]{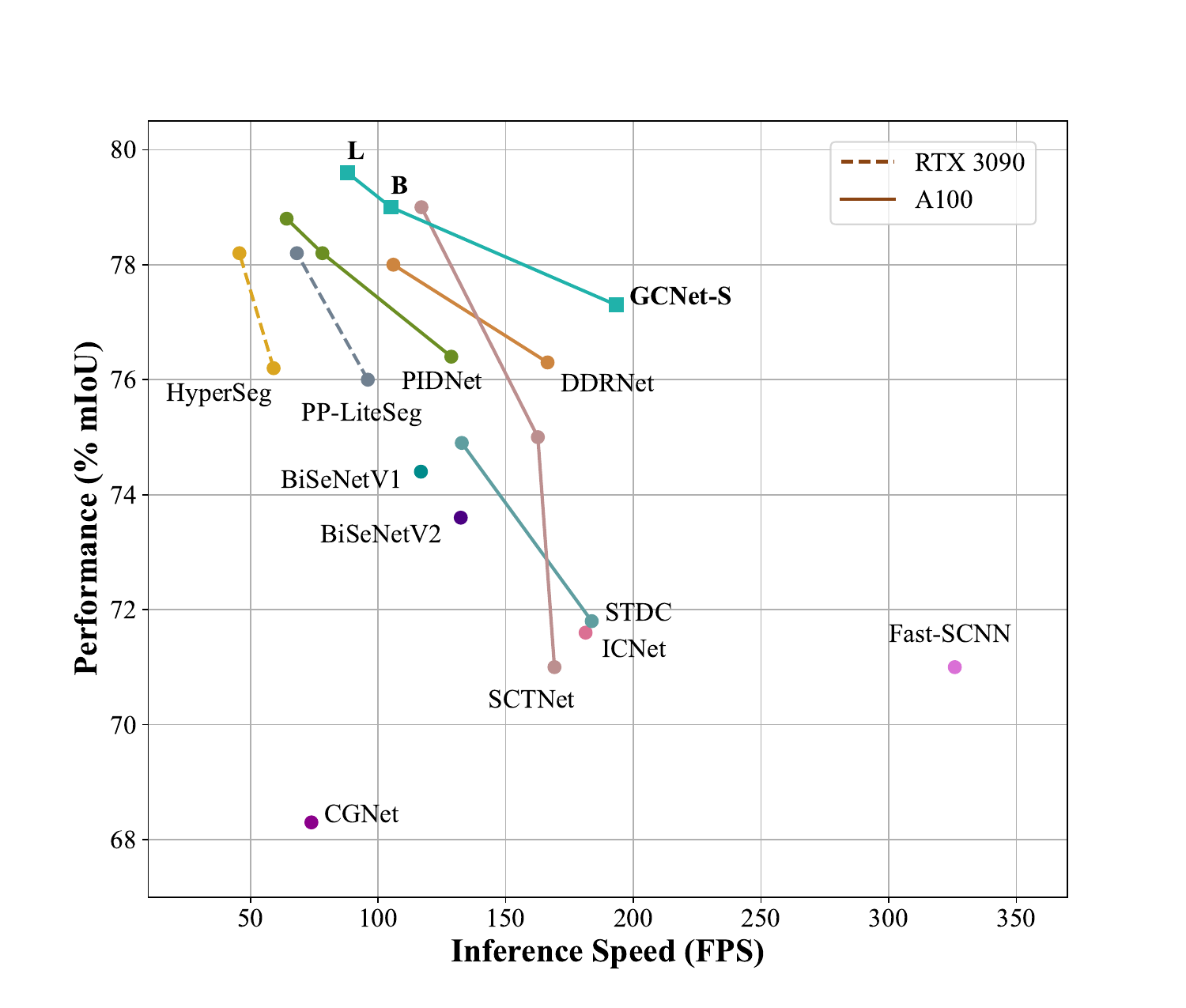}
\caption{The trade-off between inference speed and performance for real-time semantic segmentation models on the Cityscapes validation set.}
\label{0}
\end{figure}

To meet the demands of real-time applications, numerous semantic segmentation models have emerged in recent years, emphasizing both performance and inference speed. One of the earliest, ERFNet~\cite{romera2017erfnet}, reduced parameters and computational load by redesigning ResNet~\cite{he2016deep} blocks using 1D convolutional kernels and skip connections. However, its single-branch architecture limits its ability to learn spatial information from high-resolution features. In response, BiSeNetV1~\cite{yu2018bisenet}, BiSeNetV2~\cite{yu2021bisenet}, and DDRNet~\cite{pan2022deep} introduced two-branch architectures, with one branch focusing on spatial details and the other on deep semantic information. PIDNet~\cite{xu2023pidnet} expanded this idea with a three-branch design, adding a branch for boundary information. Although multi-branch models have proven effective for enhancing spatial detail capture, SCTNet~\cite{xu2024sctnet} takes a different approach by proposing a single-branch model that improves semantic representation through knowledge distillation from a high-performance transformer segmentation model~\cite{xie2021segformer}.

\begin{figure}[t]
\centering
\includegraphics[width=0.99\linewidth]{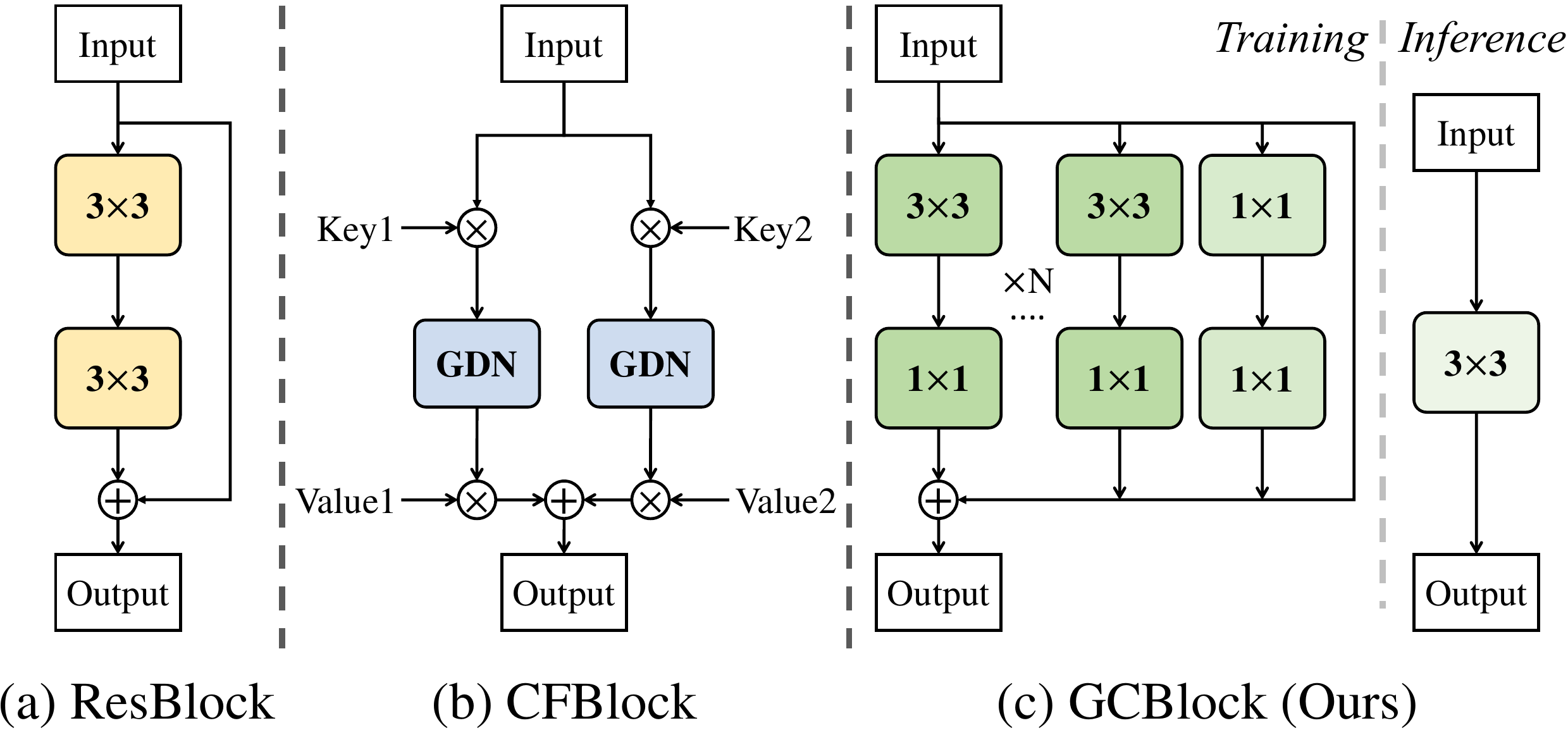}
\caption{A comparison of the proposed GCBlock with multi-path blocks: (a) Residual Block~\cite{he2016deep}, used by model~\cite{pan2022deep, xu2023pidnet, xu2024sctnet}. (b) Conv-Former Block~\cite{xu2024sctnet}, used by model~\cite{xu2024sctnet}. (c) GCBlock, a block that is scalable in both vertical and horizontal directions.}
\label{block_contract}
\end{figure}

While models mentioned above~\cite{pan2022deep, xu2023pidnet, xu2024sctnet} have achieved impressive speed and performance, their use of Residual Blocks~\cite{he2016deep} and Conv-Former Blocks~\cite{xu2024sctnet} can hinder inference speed. As shown in Figure~\ref{block_contract}, residual connection increase memory access frequency, and the complexity of Conv-Former Block, similar to transformer architectures, also affects memory access and inference efficiency. In contrast, single-path blocks~\cite{simonyan2014very} are better suited for real-time segmentation due to their unidirectional structure, which allows for faster processing and reduced memory overhead. Additionally, although SCTNet uses a single-branch architecture for inference, it relies on a high-performance teacher model during training, which raises costs. The necessity of pre-trained weights from the teacher model on a specific dataset further complicates training and increases resource demands.

Based on the aforementioned description, two key issues currently exist: (1) the decrease in inference speed caused by multi-path blocks and transformer-like convolution structures, and (2) the reliance on high-performance teacher models during training. To address issue (1), we introduce the Golden Cudgel Block (GCBlock), which employs a configuration of vertical multi-convolutions and horizontal multi-paths during training. During inference, these convolutions and paths are reparameterized into a single 3 $\times$ 3 convolution. This design seeks to simultaneously leverage the training advantages of multi-path blocks, such as avoiding issues of gradient vanishing and explosion, and the inference advantages of single-path blocks. To solve issue (2), we propose the Golden Cudgel Network (GCNet), which stacks multiple GCBlocks. During training, GCNet enlarges itself to enhance its learning capacity, while during inference, it contracts to improve speed. From a broader perspective, GCNet can effectively transform itself into a ``teacher model" without the need for an external one.

To evaluate the performance and speed of the proposed GCNet, we conducted experiments on three datasets: Cityscapes~\cite{cordts2016cityscapes}, CamVid~\cite{brostow2009semantic}, and Pascal VOC 2012~\cite{everingham2010pascal}. The results indicate that GCNet achieves a superior balance between segmentation performance and inference speed. As shown in Figure~\ref{0}, compared to other state-of-the-art real-time semantic segmentation models, GCNet demonstrates enhanced performance and faster speed. The main contributions of this paper are as follows:
\begin{itemize}
\item We propose the Golden Cudgel Network family, a model designed to enhance performance through self-enlargement and increase inference speed through self-contraction without any loss in performance. 
\item We introduce the Golden Cudgel Block, which leverages the training advantages of multi-path blocks and the inference advantages of single-path blocks through reparameterization.
\item Experiments conducted on three public datasets demonstrate that the proposed GCNet achieves a superior balance between performance and speed compared to other state-of-the-art real-time semantic segmentation models. In particular, without ImageNet pre-training, GCNet reaches 77.3\% mIoU and 193.3 FPS on the Cityscapes.
\end{itemize}
\section{Related Work}
\label{sec:related_work}

\subsection{High-performance Semantic Segmentation}

High-performance semantic segmentation models primarily focus on segmentation quality and accuracy, often at the expense of inference speed. FCN~\cite{shelhamer2017fully}, one of the earliest deep learning-based models, achieved pixel-level segmentation by replacing fully connected layers in CNNs with convolutional layers. However, its heavy downsampling results in a loss of spatial detail. To mitigate this, the DeepLab series~\cite{chen2017deeplab, chen2017rethinking, chen2018encoder} integrated dilated convolutions~\cite{YuKoltun2016} with varying rates to enhance the receptive field while preserving spatial resolution. Recently, transformer-based segmentation models~\cite{xie2021segformer, cheng2021per, cheng2022masked} have emerged, leveraging self-attention to capture long-range dependencies and improve performance. Despite their effectiveness, these models remain unoptimized for inference speed, limiting their suitability for real-time applications.

\begin{figure*}[htbp]
\centering
\includegraphics[width=0.99\linewidth]{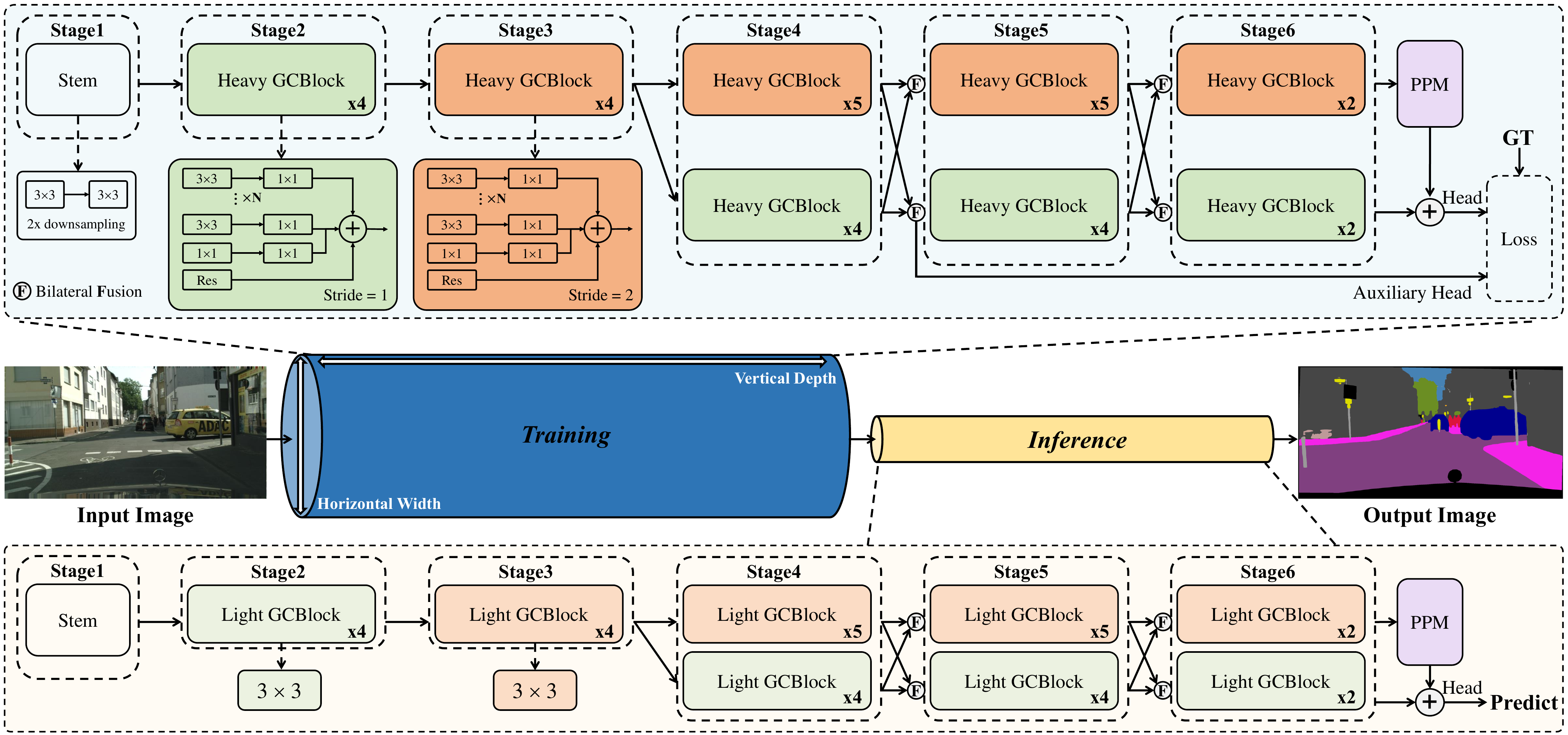}
\caption{The overall architecture of GCNet. After feature flow into two branches, the upper branch corresponds to the semantic branch, while the lower branch corresponds to the detail branch. The orange box indicates that the first block within the GCBlocks has a stride of 2, while the remaining blocks have a stride of 1. The green box signifies that all GCBlocks maintain a stride of 1. PPM refers to the Deep Aggregation Pyramid Pooling Module~\cite{pan2022deep}.}
\label{gcnet}
\end{figure*}

\subsection{Real-time Semantic Segmentation}

Real-time semantic segmentation models aim to simultaneously enhance speed and performance. However, optimizing speed through adjustments to depth, width, or structure can compromise performance, necessitating strategies to balance both. Architecturally, these models fall into two categories: single-branch and multi-branch architectures.

\textbf{Single-branch Architecture:} As single-branch architecture models, they~\cite{romera2017erfnet, wu2020cgnet, peng2022pp} typically balance performance and speed by optimizing backbone blocks or designing lightweight segmentation heads. However, with the rise of multi-branch models that offer enhanced performance and speed, single-branch models face challenges. Despite this, numerous studies continues to advance single-branch models. For example, STDC~\cite{fan2021rethinking} addresses the speed slow addition of branches in BiSeNet~\cite{yu2018bisenet} by introducing the Detail Aggregation Module, which retains spatial information through a single-branch approach. SCTNet~\cite{xu2024sctnet} utilizes knowledge distillation with SegFormer~\cite{xie2021segformer} as the teacher model, employing a Conv-Former Block to bridge the semantic gap between CNN and transformer features, allowing for more effective learning of rich semantic information.

\textbf{Multi-branch Architecture:} Single-branch architecture models use skip connections to link feature maps between the backbone and head, preserving spatial details. In contrast, multi-branch models add branches to learn both spatial details and boundary information. BiSeNetV1~\cite{yu2018bisenet} and V2~\cite{yu2021bisenet} propose a two-branch architecture where one branch focuses on deep semantic information and the other on spatial details, with no weight sharing between them. Conversely, models like Fast-SCNN~\cite{poudel2019fast} and DDRNet~\cite{pan2022deep} share some low-level backbone weights. Fast-SCNN extracts shallow features and splits them into two branches for feature retention and global feature extraction, while DDRNet performs multiple bilateral fusions to efficiently merge information. PIDNet~\cite{xu2023pidnet} introduces a three-branch architecture to capture spatial detail, deep semantic information, and boundary information.

Whether single-branch models or multi-branch models, although they continually compete in architectural design to optimize performance and speed, their inference speeds are limited by multi-path blocks, as shown in Figure~\ref{block_contract}. In contrast, our GCNet, based on a two-branch architecture, enhances its learning capability during training by vertically increasing convolutions and horizontally expanding paths. During inference, it improves inference speed by reparameterizing multi-convolution multi-path blocks into a single convolution, all without sacrificing performance. Although some studies~\cite{ding2019acnet, ding2021diverse, ding2021repvgg} have employed similar methods to enhance model performance and speed in image classification tasks, such approaches have been seldom explored in semantic segmentation tasks.

\section{Methodology}
\label{sec:method}

In this section, we first present the overall framework of GCNet. Then, we describe the training structure of the GCBlock and its reparameterization into a single convolution during inference. Finally, we detail the supervisory strategy and loss function used in model training.

\subsection{Golden Cudgel Network}

To address the issue of increased training costs associated with the use of teacher models, we propose the Golden Cudgel Network (GCNet). We establish a two-branch architecture as baseline model and subsequently enhance it by deepening the model with additional convolution and broadening it with multiple paths, thereby augmenting its learning capacity. After training, we reduce the model's convolution and the number of paths to improve inference speed, a process that does not compromise performance. From a macro perspective, the self-enlarged GCNet functions as the ``teacher model", while the self-contracted GCNet serves as the ``student model".

\begin{table}[t]
\renewcommand{\arraystretch}{1.4}
\caption{Details of the different versions of GCNet. H represents height, W represents width, and C represents channels. For GCNet-S, C is set to 32, while for GCNet-M and GCNet-L, C is set to 64. In rows $S_{4}$, $S_{5}$, and $S_{6}$, the left side of each column indicates the dimensions of the semantic branch, while the right side indicates the dimensions of the detail branch.}
\centering
\resizebox{\linewidth}{!}
{
\begin{tabular}{c|c|c|c} 
\specialrule{0.2em}{0em}{0.3em}
\multirow{2}*{\textbf{Stage}}           & \multirow{2}*{\textbf{Output Dimension}}                                        & \multicolumn{2}{c}{\textbf{Quantity of Blocks}}                                                \\ \cline{3-4} 
                                        &                                                                                 & \textbf{GCNet-S/M}                                 & \textbf{GCNet-L}                          \\
\specialrule{0.05em}{0em}{0em}
Input                                   & H $\times$ W $\times$ 3                                                         &                                                    &                                           \\
\specialrule{0.05em}{0em}{0em}
$S_{1}$                                 & H/4 $\times$ W/4 $\times$ C                                                     &                                                    &                                           \\
\specialrule{0.05em}{0em}{0em}
$S_{2}$                                 & H/4 $\times$ W/4 $\times$ C                                                     & 4                                                  & 5                                         \\
\specialrule{0.05em}{0em}{0em}
$S_{3}$                                 & H/8 $\times$ W/8 $\times$ 2C                                                    & 4                                                  & 5                                         \\
\specialrule{0.05em}{0em}{0em}
$S_{4}$                                 & H/16 $\times$ W/16 $\times$ 4C, H/8 $\times$ W/8 $\times$ 2C                    & 5, 4                                               & 5, 5                                      \\
\specialrule{0.05em}{0em}{0em}
$S_{5}$                                 & H/32 $\times$ W/32 $\times$ 8C, H/8 $\times$ W/8 $\times$ 2C                    & 5, 4                                               & 5, 5                                      \\
\specialrule{0.05em}{0em}{0em}
$S_{6}$                                 & H/64 $\times$ W/64 $\times$ 16C, H/8 $\times$ W/8 $\times$ 4C                   & 2, 2                                               & 3, 3                                      \\
\specialrule{0.2em}{0.3em}{0em}
\end{tabular}
}
\label{tab_architecture}
\end{table}

As shown in Figure~\ref{gcnet}, GCNet is a two-branch architecture model, with its backbone composed of stacked GCBlocks. After extracting shallow features (stage 3), the features flow into two branches: the semantic branch and the detail branch. The semantic branch is designed to learn deep semantic information, while the detail branch focuses on learning spatial detail information. After passing through stage 4 or stage 5, the features from the semantic branch are fused with those from the detail branch after channel compression and upsampling. Conversely, the features from the detail branch undergo channel expansion and downsampling before being integrated with the features from the semantic branch. This process enhances the richness of the features from both branches. We do not utilize various attention modules~\cite{hu2018squeeze, fu2019dual, wang2020eca, xu2023pidnet} for feature fusion, as this would complicate the model and decrease inference speed. Instead, we utilize 3 $\times$ 3 convolutions for channel compression and expansion, along with bilinear interpolation for upsampling. After stage 6, the features from the semantic branch are processed through a pyramid pooling module~\cite{pan2022deep} to generate richer feature representations, which are then upsampled and fused with the features from the detail branch before being passed to the segmentation head for prediction. The segmentation head of GCNet consists of a 3 $\times$ 3 convolution followed by a 1 $\times$ 1 convolution. The 3 $\times$ 3 convolution is used to integrate the features from both branches while adjusting the channel dimension to $O_{c}$, and the 1 $\times$ 1 convolution is used to align the number of classes. For instance, if the number of classes is 19, the 1 $\times$ 1 convolution will adjust $O_{c}$ to 19. Additionally, we implement an auxiliary segmentation head during training to further enhance GCNet's performance.

As illustrated in Figure~\ref{gcnet}, the backbone of GCNet consists of six stages. Stage 1 serves as the Stem, comprising two 3 $\times$ 3 convolutions with a stride of 2, which is intended for rapid downsampling of the image. The subsequent stages are made up of stacked GCBlocks. Notably, the orange box in the figure indicates that the first block within the GCBlocks has a stride of 2, while the remaining blocks have a stride of 1; the green box denotes that all GCBlocks maintain a stride of 1. To accommodate various application scenarios, we adjusted both the quantity of stacked GCBlocks and the convolution width of the GCBlocks at each stage, resulting in three distinct versions of GCNet: GCNet-S, GCNet-M, and GCNet-L, as detailed in Table~\ref{tab_architecture}. The OC of the segmentation heads for GCNet-S, GCNet-M, and GCNet-L are 64, 128, and 256, respectively.

\subsection{Golden Cudgel Block}

\noindent\textbf{Structure.} To address speed degradation from the complex structures of multi-path and transformer-like convolution blocks, we propose the Golden Cudgel Block (GCBlock). During training, GCBlock expands into a multi-convolution, multi-path structure to leverage the benefits of these blocks. During inference, it simplifies to a single convolution through reparameterization, enhancing efficiency. Specifically, based on the bottleneck structure~\cite{he2016deep}, we achieve vertical reparameterization of convolutions into a single 3 $\times$ 3 convolution by removing activation functions between convolutional layers, keeping only one at the output. During our investigation, we found that after training, the reduction in model parameter values hindered the lossless fusion between the first 1 $\times$ 1 convolution and the 3 $\times$ 3 convolution within the bottleneck. Thus, we eliminated the first 1 $\times$ 1 convolution, retaining one 3 $\times$ 3 convolution and one 1 $\times$ 1 convolution. Additionally, we introduced Path$\mathbf{_{1\times1\_1\times1}}$ (consisting of two 1 $\times$ 1 convolutions) and more Path$\mathbf{_{3\times3\_1\times1}}$ (comprising one 3 $\times$ 3 and one 1 $\times$ 1 convolution) to enhance learning capability. Although GCBlock contains multiple convolutions and paths, these components integrate into a single 3 $\times$ 3 convolution during inference, ensuring both performance and speed. The structure of GCBlock are illustrated in Figure~\ref{gcnet}, where N indicating the number of Path$\mathbf{_{3\times3\_1\times1}}$. We will detail the process of vertically reparameterizing each path into a single 3 $\times$ 3 convolution and then summing them horizontally.

\noindent\textbf{Conv-Bn to Conv.} To enhance the stability of training, Batch Normalization (BN) is typically applied after convolution to perform channel normalization and linear scaling. Before reparameterizing between convolutions, it is necessary to merge the convolution and BN. Let $\mu$, $\sigma$, $\gamma$, and $\beta \in \mathbb{R}^{C_{out}}$, represent the mean, variance, scaling factor, and bias in BN, respectively, and let $W \in \mathbb{R}^{C_{out} \times C_{in} \times k \times k}$ and $B \in \mathbb{R}^{C_{out}}$ denote the weight and bias of a k $\times$ k convolution. $C_{in}$ represents the number of input channels, $C_{out}$ and represents the number of output channels. The merging of the convolution and BN can be computed as follows:
\begin{equation}
\label{eq1}
W^{\prime} = \frac{\gamma}{\sqrt{\sigma + \varepsilon}} W, B^{\prime} = \frac{(B - \mu)\gamma}{\sqrt{\sigma + \varepsilon}} + \beta,
\end{equation}
where $\varepsilon$ is a constant term used in BN to prevent division by zero, and it is typically set to 1e-5.

\noindent\textbf{Path$\mathbf{_{3\times3\_1\times1}}$ to Conv$\mathbf{_{3\times3}}$.} For a concatenation consisting of a 3 $\times$ 3 convolution followed by a 1 $\times$ 1 convolution, if there are no batch normalization layers or activation functions in between, they can be merged into a single 3 $\times$ 3 convolution. Let the weight and bias of the 3 $\times$ 3 convolution be denoted as $W_{3 \times 3} \in \mathbb{R}^{C_{out1} \times (C_{in} \times 3 \times 3)}$ and $B_{3 \times 3} \in \mathbb{R}^{C_{out1}}$, respectively, and the weight and bias of the 1 $\times$ 1 convolution be denoted as $W_{1 \times 1} \in \mathbb{R}^{C_{out2} \times (C_{out1} \times 1 \times 1)}$ and $B_{1 \times 1} \in \mathbb{R}^{C_{out2}}$. Given an input $x$, the output $y$ can be computed as follows:
\begin{align}
\label{eq2}
y &= W_{1 \times 1} * (W_{3 \times 3} * x + B_{3 \times 3}) + B_{1 \times 1}                                                   \\
\notag
  &= W_{1 \times 1} \cdot I_{2}(W_{3 \times 3} \cdot I_{1}(x) + B_{3 \times 3}) + B_{1 \times 1}                               \\
\notag
  &= W_{1 \times 1} \cdot I_{2}(W_{3 \times 3} \cdot I_{1}(x)) + W_{1 \times 1} \cdot I_{2}(B_{3 \times 3}) + B_{1 \times 1}   \\
\notag
  &= W_{1 \times 1} \cdot I_{2}(W_{3 \times 3} \cdot I_{1}(x)) + W_{1 \times 1} \cdot B_{3 \times 3} + B_{1 \times 1},         
\notag
\end{align}
where $*$ denotes the convolution operation, while $\cdot$ represents matrix multiplication. To perform matrix operations, $I(\cdot)$ denotes the im2col operator, which is utilized to transform the input $x$ into a two-dimensional matrix based on the shape of the convolution weight. For instance, $I(\cdot)$ will convert the input $x$ with dimensions $C_{in} \times H \times W$ into a matrix of shape $(C_{in} \times 3 \times 3) \times (H' \times W')$, according to the dimensions of $W_{3 \times 3}$. Since $B_{3 \times 3}$ is a constant sequence, $W_{1 \times 1} \cdot I_{2}(B_{3 \times 3})$ is equivalent to $W_{1 \times 1} \cdot B_{3 \times 3}$. Furthermore, since the shape of $W_{1 \times 1}$ is $C_{out2} \times (C_{out1} \times 1 \times 1)$ and the shape of $W_{3 \times 3} \cdot I_{1}(x)$ is $C_{out1} \times (H' \times W')$, we can conclude that:
\begin{equation}
\label{eq3}
\begin{split}
&~~~~~ W_{1 \times 1} \cdot I_{2}(W_{3 \times 3} \cdot I_{1}(x)) ~~~~~  \\
&= W_{1 \times 1} \cdot W_{3 \times 3} \cdot I_{1}(x).
\end{split}
\end{equation}
Subsequently, by substituting Equation~\ref{eq3} into Equation~\ref{eq2}, the following result is obtained:
\begin{align}
\label{eq4}
\notag
y &= W_{1 \times 1} \cdot I_{2}(W_{3 \times 3} \cdot I_{1}(x)) + W_{1 \times 1} \cdot B_{3 \times 3} + B_{1 \times 1}       \\
\notag
  &= W_{1 \times 1} \cdot W_{3 \times 3} \cdot I_{1}(x) + W_{1 \times 1} \cdot B_{3 \times 3} + B_{1 \times 1}              \\
\notag
  &= (W_{1 \times 1} \cdot W_{3 \times 3}) \cdot I_{1}(x) + (W_{1 \times 1} \cdot B_{3 \times 3} + B_{1 \times 1})          \\
  &= W'_{3 \times 3} \cdot I_{1}(x) + B'_{3 \times 3},
\end{align}
where $W'_{3 \times 3}$ represents the new weight, while $B'_{3 \times 3}$ represents the new bias. It is evident that a concatenation consisting of a 3 $\times$ 3 convolution followed by a 1 $\times$ 1 convolution can be merged into a new 3 $\times$ 3 convolution without any loss in performance.

\noindent\textbf{Path$\mathbf{_{1\times1\_1\times1}}$ to Conv$\mathbf{_{3\times3}}$.} A concatenation of two 1 $\times$ 1 convolutions can also be merged into a 3 $\times$ 3 convolution according to Equation~\ref{eq4}. The reason is that the first 1 $\times$ 1 convolution can be viewed as a 3 $\times$ 3 convolution with a non-zero weight value at the center and zero weight values in the surrounding positions. Notably, to satisfy the conditions in Equation~\ref{eq4}, the stride of the first 1 $\times$ 1 convolution is set to 2 during downsampling, rather than being applied to the subsequent 1 $\times$ 1 convolution. Theoretically, the Path$\mathbf{_{1\times1\_1\times1}}$ can accommodate an arbitrary number of 1 $\times$ 1 convolutions; however, experiments have shown that stacking two 1 $\times$ 1 convolutions yields the best model performance.

\noindent\textbf{Path$\mathbf{_{residual}}$ to Conv$\mathbf{_{3\times3}}$.} For residual connection, we first construct a 1 $\times$ 1 convolution with weight $W_{residual} \in \mathbb{R}^{C_{out} \times (C_{in} \times 1 \times 1)}$ ($C_{out} = C_{in}$) and bias $B_{residual} \in \mathbb{R}^{C_{out}}$ (set to 0) to equivalently replace it. Next, this 1 $\times$ 1 convolution is transformed into a 3 $\times$ 3 convolution. Specifically, we traverse $W_{residual}$ using $C_{out}$ and $C_{in}$. When $c_{out}$ equals $c_{in}$ during the traversal, the weight value at that position is set to 1; otherwise, it is set to 0. After completing the traversal, the resulting 1 $\times$ 1 convolution is equivalent to the residual connection. Since a 1 $\times$ 1 convolution is a special case of a 3 $\times$ 3 convolution, it is straightforward to convert the 1 $\times$ 1 convolution into a 3 $\times$ 3 convolution. In Path$\mathbf{_{residual}}$, we use a BN layer, so it is also necessary to integrate the converted convolution with the BN. Notably, when the stride of the GCBlock is set to 2, the residual connection is not used.

\noindent\textbf{Multipath to Single Convolution.} After reparameterizing each path into 3 $\times$ 3 convolutions, parallel sets of $n$ weights $W \in \mathbb{R}^{C_{out} \times (C_{in} \times 3 \times 3)}$ and $n$ biases $B \in \mathbb{R}^{C_{out}}$ are obtained. Let $W_{i}$ represent the $i$-th weight and $B_{i}$ the $i$-th bias, where $i$ is less than or equal to $n$. Given an input $x$, the output $y$ from multiple paths is calculated as:
\begin{equation}
\label{eq5}
\begin{split}
y &= \sum_{i=1}^{n}\Big(W_{i} * x + B_{i}\Big)                      \\
  &= \sum_{i=1}^{n}\Big(W_{i} \cdot I_{i}(x) + B_{i}\Big)           \\
  &= \Big(\sum_{i=1}^{n}W_{i}\Big) \cdot I(x) + \sum_{i=1}^{n}B_{i} \\
  &= W' * x + B'
\end{split}
\end{equation}
where $I_{i}$ is the $i$-th im2col operator. Since the shape of $W_{i}$ is consistent, all $I_{i}$ are the same. This allows for direct summation of these 3 $\times$ 3 convolutions, enabling the acquisition of a new 3 $\times$ 3 convolution without performance loss.

\subsection{Deep Supervision and Loss Function}

Previous research~\cite{yu2021bisenet, pan2022deep, xu2023pidnet} has demonstrated that adding an auxiliary segmentation head to appropriate positions in the model during training can improve the model's performance without increasing inference time. Building on prior work, we introduced an auxiliary segmentation head in GCNet, which is removed during inference. As illustrated in Figure~\ref{gcnet}, after the bilateral fusion in the stage 4, the features from the detail branch are passed to the auxiliary segmentation head for loss calculation. Notably, the structure of the auxiliary segmentation head is identical to that of the primary segmentation head. Based on the losses from both the segmentation head and the auxiliary segmentation head, the total loss can be expressed as: 
\begin{equation}
\label{eq6}
L = L_{sh} + {\alpha}L_{ash},
\end{equation}
where $L_{sh}$ represents the loss from the segmentation head, $L_{ash}$ denotes the loss from the auxiliary segmentation head, and $\alpha$ is the weight coefficient, which is set to 0.4 in GCNet. To effectively handle imbalanced data and difficult samples, we employed OHEM Cross Entropy~\cite{shrivastava2016training} as the loss function, consistent with previous work~\cite{xu2023pidnet}.

\begin{table}[t]
\renewcommand{\arraystretch}{1.1}
\setlength{\tabcolsep}{24pt}
\caption{Ablation study on Path$\mathbf{_{1\times1\_1\times1}}$ for GCNet-S. ``Number" indicates the number of convolutions used in the path, ``Memory" refers to the GPU memory utilized during training, and ``Time" represents the training duration in hours.}
\centering
\resizebox{\linewidth}{!}
{
\begin{tabular}{cccc} 
\specialrule{0.2em}{0em}{0.3em}
\textbf{Number}          & \textbf{Memory}           & \textbf{Time}             & \textbf{mIoU} \\ 
\specialrule{0.12em}{0.3em}{0.3em}
0                        & 20.58 GiB                 & 4.0 h                     & 76.1          \\ 
1                        & 21.87 GiB                 & 4.5 h                     & 76.6          \\ 
2                        & 24.61 GiB                 & 5.0 h                     & 76.7          \\
3                        & 27.31 GiB                 & 5.4 h                     & 76.4          \\
\specialrule{0.2em}{0.3em}{0em}
\end{tabular}
}
\label{ablation_path1x1}
\end{table}

\section{Experients}
\label{sec:experients}

\subsection{Datasets and Implementation Details}

To demonstrate the effectiveness of GCNet, we conducted experiments on three public datasets: Cityscapes~\cite{cordts2016cityscapes}, CamVid~\cite{brostow2009semantic}, and Pascal VOC 2012~\cite{everingham2010pascal}. During training on Cityscapes, we did not utilize ImageNet~\cite{deng2009imagenet} pretrained weight and instead trained from scratch using four A100 GPUs. For training on CamVid and Pascal VOC 2012, we fine-tuned the model using weight from Cityscapes. During inference, we used a single A100 GPU. More details regarding the datasets, training settings, inference settings, and evaluation metrics can be found in Appendix~\ref{More Experimental Details}.

\begin{figure}[htbp]
\centering
\includegraphics[width=0.99\linewidth]{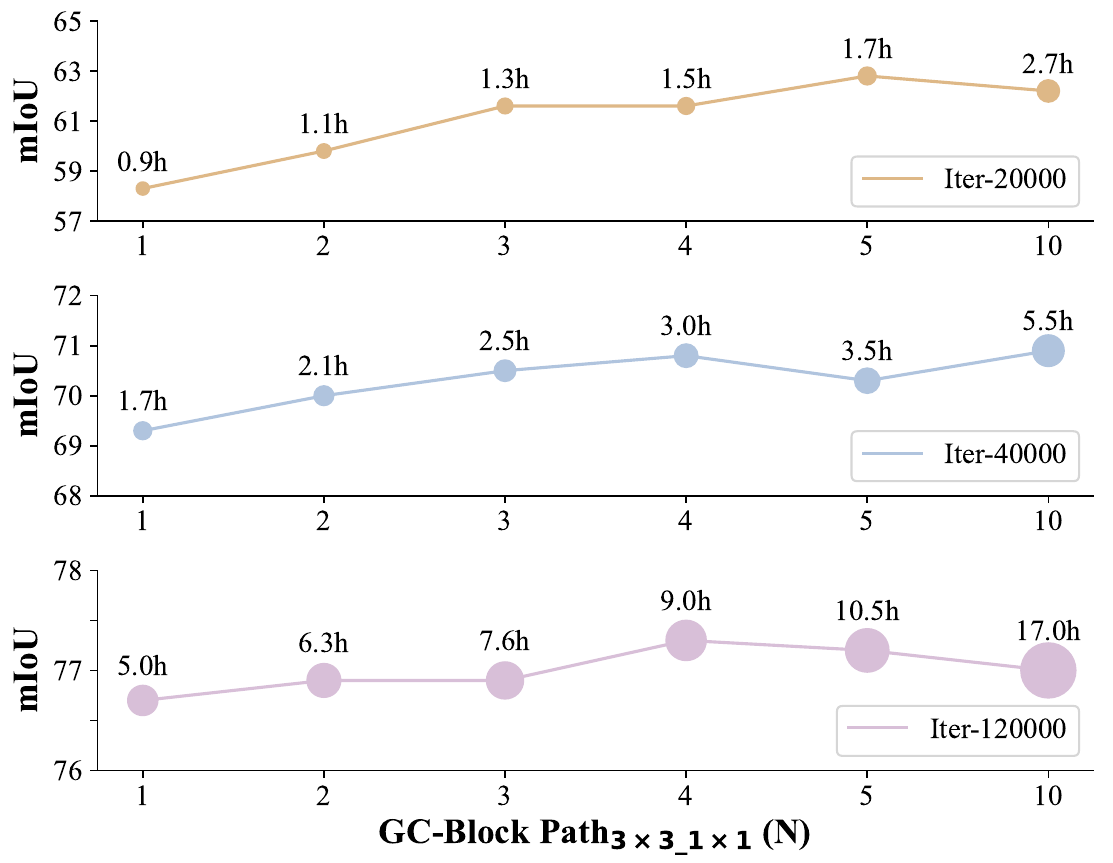}
\caption{Ablation study on Path$\mathbf{_{3\times3\_1\times1}}$ for GCNet-S. ``N" indicates the number of Path$\mathbf{_{3\times3\_1\times1}}$, while ``Iter-20000" signifies that 20000 iterations were completed, and so on.}
\label{ablationN}
\end{figure}

\begin{figure}[b]
\centering
\includegraphics[width=\linewidth]{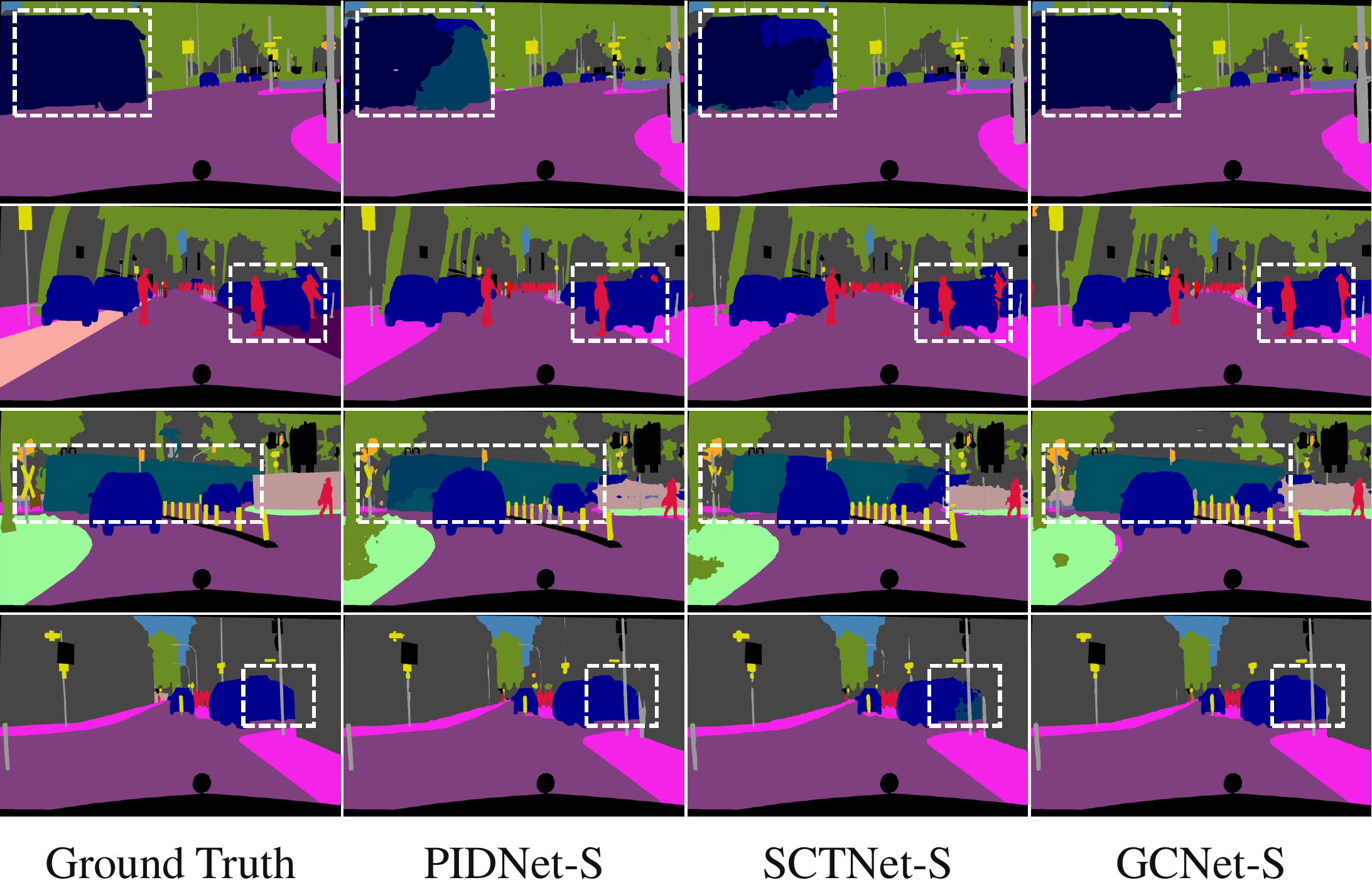}
\caption{Visualization of different segmentation models on the Cityscapes validation set.}
\label{vis_seg}
\end{figure}

\begin{table*}[htbp]
\renewcommand{\arraystretch}{1.12}
\caption{Comparison on the Cityscapes validation set. ``*” indicates that the model was reimplemented and retested by us (with convolution and batch normalization fused). ``ImageNet" indicates whether the model utilizes ImageNet pretrained weight.}
\centering
\resizebox{\linewidth}{!}
{
\begin{tabular}{lccccccc} 
\specialrule{0.2em}{0em}{0.3em}
\multirow{2}*{\textbf{Method}}              & \multirow{2}*{\textbf{Resolution}}   & \multirow{2}*{\textbf{GPU}}  & \multirow{2}*{\textbf{FPS}}  & \multirow{2}*{\textbf{Params}}   & \multirow{2}*{\textbf{GFLOPs}}   & \multirow{2}*{\textbf{ImageNet}}   & \multirow{2}*{\textbf{mIoU (\%)}}    \\ 
                                            &                                      &                          &                              &                                  &                                  &                                    &               \\
\specialrule{0.15em}{0.3em}{0.3em}
ICNet*~\cite{zhao2018icnet}                 & $1024 \times 2048$                   & A100                     & 181.3                        & 24.9 M                           & 74.3                             & \ding{55}                          & 71.6          \\
\specialrule{0.05em}{0.3em}{0.3em}
Fast-SCNN*~\cite{poudel2019fast}            & $1024 \times 2048$                   & A100                     & 325.9                        & 1.4 M                            & 7.3                              & \ding{55}                          & 71.0          \\
\specialrule{0.05em}{0.3em}{0.3em}
SwiftNetRN-18~\cite{orsic2019defense}       & $1024 \times 2048$                   & GTX 1080Ti               & 39.9                         & 11.8 M                           & 104.0                            & \ding{51}                          & 75.4          \\
\specialrule{0.05em}{0.3em}{0.3em}
CGNet*~\cite{wu2020cgnet}                   & $1024 \times 2048$                   & A100                     & 73.9                         & 0.5 M                            & 27.5                             & \ding{55}                          & 68.3          \\
\specialrule{0.05em}{0.3em}{0.3em}
BiSeNetV1*~\cite{yu2018bisenet}             & $1024 \times 2048$                   & A100                     & 116.8                        & 13.3 M                           & 118.0                            & \ding{51}                          & 74.4          \\
BiSeNetV2*~\cite{yu2021bisenet}             & $1024 \times 2048$                   & A100                     & 132.4                        & 3.4 M                            & 98.4                             & \ding{55}                          & 73.6          \\
\specialrule{0.05em}{0.3em}{0.3em}
STDC1*~\cite{fan2021rethinking}             & $1024 \times 2048$                   & A100                     & 183.7                        & 8.3 M                            & 67.5                             & \ding{55}                          & 71.8          \\
STDC2*~\cite{fan2021rethinking}             & $1024 \times 2048$                   & A100                     & 132.8                        & 12.3 M                           & 93.8                             & \ding{55}                          & 74.9          \\
\specialrule{0.05em}{0.3em}{0.3em}
HyperSeg-M~\cite{nirkin2021hyperseg}        & $512 \times 1024$                    & RTX 3090                 & 59.1                         & 10.1 M                           & 7.5                              & \ding{51}                          & 76.2          \\
HyperSeg-S~\cite{nirkin2021hyperseg}        & $768 \times 1536$                    & RTX 3090                 & 45.7                         & 10.2 M                           & 17.0                             & \ding{51}                          & 78.2          \\
\specialrule{0.05em}{0.3em}{0.3em}
PP-LiteSeg-T2~\cite{peng2022pp}             & $768 \times 1536$                    & RTX 3090                 & 96.0                         & -                                & -                                & \ding{51}                          & 76.0          \\
PP-LiteSeg-B2~\cite{peng2022pp}             & $768 \times 1536$                    & RTX 3090                 & 68.2                         & -                                & -                                & \ding{51}                          & 78.2          \\
\specialrule{0.05em}{0.3em}{0.3em}
DDRNet-23-Slim*~\cite{pan2022deep}          & $1024 \times 2048$                   & A100                     & 166.4                        & 5.7 M                            & 36.3                             & \ding{55}                          & 76.3          \\
DDRNet-23*~\cite{pan2022deep}               & $1024 \times 2048$                   & A100                     & 106.0                        & 20.3 M                           & 143.0                            & \ding{55}                          & 78.0          \\
\specialrule{0.05em}{0.3em}{0.3em}
PIDNet-S*~\cite{xu2023pidnet}               & $1024 \times 2048$                   & A100                     & 128.7                        & 7.7 M                            & 47.3                             & \ding{55}                          & 76.4          \\
PIDNet-M*~\cite{xu2023pidnet}               & $1024 \times 2048$                   & A100                     & 78.2                         & 28.7 M                           & 177.0                            & \ding{55}                          & 78.2          \\
PIDNet-L*~\cite{xu2023pidnet}               & $1024 \times 2048$                   & A100                     & 64.2                         & 37.3 M                           & 275.0                            & \ding{55}                          & 78.8          \\
\specialrule{0.05em}{0.3em}{0.3em}
SCTNet-S-Seg50*~\cite{xu2024sctnet}         & $512 \times 1024$                    & A100                     & 169.1                        & 4.6  M                           & 7.1                              & \ding{55}                          & 71.0          \\
SCTNet-S-Seg75*~\cite{xu2024sctnet}         & $768 \times 1536$                    & A100                     & 168.7                        & 4.6  M                           & 16.0                             & \ding{55}                          & 74.7          \\
SCTNet-B-Seg50*~\cite{xu2024sctnet}         & $512 \times 1024$                    & A100                     & 162.6                        & 17.4 M                           & 28.1                             & \ding{55}                          & 75.0          \\
SCTNet-B-Seg75*~\cite{xu2024sctnet}         & $768 \times 1536$                    & A100                     & 157.3                        & 17.4 M                           & 63.2                             & \ding{55}                          & 78.5          \\
SCTNet-B-Seg100*~\cite{xu2024sctnet}        & $1024 \times 2048$                   & A100                     & 117.0                        & 17.4 M                           & 112.3                            & \ding{55}                          & 79.0          \\
\specialrule{0.05em}{0.3em}{0.3em}
\rowcolor{green!25}
GCNet-S                                     & $1024 \times 2048$                   & A100                     & 193.3                        & 9.2 M                            & 45.2                             & \ding{55}                          & 77.3          \\
\rowcolor{green!25}
GCNet-M                                     & $1024 \times 2048$                   & A100                     & 105.0                        & 34.2 M                           & 178.0                            & \ding{55}                          & 79.0          \\
\rowcolor{green!25}
GCNet-L                                     & $1024 \times 2048$                   & A100                     & 88.0                         & 45.2 M                           & 232.0                            & \ding{55}                          & 79.6          \\
\specialrule{0.2em}{0.3em}{0em}
\end{tabular}
}
\label{tab_cityscapes}
\end{table*}

\begin{table}[t]
\renewcommand{\arraystretch}{1.12}
\caption{Comparison on the CamVid test set and the Pascal VOC 2012 validation set. All models were reimplemented by us and retested on the A100. The inference resolution for CamVid was set at $720 \times 960$, while for VOC it was set at $512 \times 2048$. $\dagger$ indicates that ImageNet pretrained weight were used.}
\centering
\resizebox{\linewidth}{!}
{
\begin{tabular}{lcccc} 
\specialrule{0.2em}{0em}{0.3em}
\multirow{2}*{\textbf{Method}}                    & \multicolumn{2}{c}{\textbf{CamVid}}                               & \multicolumn{2}{c}{\textbf{VOC}}                                      \\ \cmidrule(lr){2-3} \cmidrule(lr){4-5}
                                                  & \textbf{FPS}                  & \textbf{mIoU (\%)}                & \textbf{FPS}                  & \textbf{mIoU (\%)}                    \\
\specialrule{0.12em}{0.3em}{0.3em}
CGNet~\cite{wu2020cgnet}                          & 102.6                         & 69.4                              & 104.7                         & 42.2                                  \\
\specialrule{0.04em}{0.3em}{0.3em}
BiSeNetV1$^{\dagger}$~\cite{yu2018bisenet}        & 276.0                         & 73.3                              & 214.3                         & 56.0                                  \\
BiSeNetV2~\cite{yu2021bisenet}                    & 240.6                         & 75.8                              & 228.6                         & 50.0                                  \\
\specialrule{0.04em}{0.3em}{0.3em}
STDC1~\cite{fan2021rethinking}                    & 240.8                         & 70.2                              & 227.0                         & 52.9                                  \\
STDC2~\cite{fan2021rethinking}                    & 149.8                         & 71.4                              & 137.2                         & 56.3                                  \\
\specialrule{0.04em}{0.3em}{0.3em}
DDRNet-23-Slim~\cite{pan2022deep}                 & 189.3                         & 76.2                              & 174.6                         & 54.2                                  \\
DDRNet-23~\cite{pan2022deep}                      & 171.1                         & 78.2                              & 165.4                         & 55.8                                  \\
\specialrule{0.04em}{0.3em}{0.3em}
PIDNet-S~\cite{xu2023pidnet}                      & 142.8                         & 77.2                              & 137.4                         & 54.0                                  \\
PIDNet-M~\cite{xu2023pidnet}                      & 128.9                         & 78.6                              & 123.1                         & 57.5                                  \\
PIDNet-L~\cite{xu2023pidnet}                      & 108.4                         & 78.7                              & 104.0                         & 58.2                                  \\
\specialrule{0.04em}{0.3em}{0.3em}
\rowcolor{green!25}
GCNet-S                                           & 210.6                         & 76.6                              & 196.9                         & 54.9                                  \\
\rowcolor{green!25}
GCNet-M                                           & 190.0                         & 78.5                              & 173.4                         & 58.2                                  \\
\rowcolor{green!25}
GCNet-L                                           & 170.4                         & 79.1                              & 145.1                         & 58.8                                  \\
\specialrule{0.2em}{0.3em}{0em}
\end{tabular}
}
\label{tab_camvid_voc}
\end{table}

\subsection{Ablation Studies}

\noindent\textbf{Number of Convolutions in Path$\mathbf{_{1\times1\_1\times1}}$.} To validate the effectiveness of Path$\mathbf{_{1\times1\_1\times1}}$, we conducted an ablation study by varying the number of 1 $\times$ 1 convolutions in this path. Specifically, we established a baseline consisting of a Path$\mathbf{_{3\times3\_1\times1}}$ and a Path$\mathbf{_{residual}}$, and then incrementally increased the number of 1 $\times$ 1 convolutions in Path$\mathbf{_{1\times1\_1\times1}}$ from 0 to 4. As shown in Table~\ref{ablation_path1x1}, both the memory usage and training time increased with the number of convolutions, calculated based on four A100 GPUs. At 0 convolutions (baseline), the model only achieved 76.1\% mIoU. However, as the number of convolutions increased, the mIoU improved, reaching peak performance with two convolutions. Consequently, we set the number of 1 $\times$ 1 convolutions in this path to 2.

\noindent\textbf{More Path$\mathbf{_{3\times3\_1\times1}}$.} To validate the effectiveness of Path$\mathbf{_{3\times3\_1\times1}}$, we conducted an ablation study by varying the number of this path. The baseline comprised a Path$\mathbf{_{3\times3\_1\times1}}$, a Path$\mathbf{_{1\times1\_1\times1}}$, and a Path$\mathbf{_{residual}}$. Although theoretically the number of Path$\mathbf{_{3\times3\_1\times1}}$ can be increased indefinitely, extensive research indicates that overly wide or deep models do not necessarily lead to improved performance; our experiments corroborate this finding. As shown in Figure~\ref{ablationN}, the model's mIoU begins to increase with the addition of Path$\mathbf{_{3\times3\_1\times1}}$, peaking at N=4 or N=5. However, at N=10, the model's performance deteriorates, with training time increasing by over 80\% compared to N=4. Additionally, we observed that with fewer iterations, increasing the number of Path$\mathbf{_{3\times3\_1\times1}}$ can significantly enhance performance (a 4.5\% increase at 20000 iterations). Considering the training cost and model performance, we set different values of N for the various versions of GCNet: N=4 for GCNet-S, N=2 for GCNet-M, and N=2 for GCNet-L.

\subsection{Comparison with State-of-the-art Methods}

\noindent\textbf{Cityscapes.} The experimental results on the Cityscapes dataset are shown in Table~\ref{tab_cityscapes}. To ensure a fair comparison, we made every effort to reproduce each model on the same hardware and tested their inference speeds. The results indicate that our GCNet achieves a better balance between performance and speed. Specifically, GCNet-S achieves 77.3\% mIoU while reaching 193.3 FPS, significantly outperforming other models of similar scale. GCNet-M also performs admirably, surpassing both PIDNet-M and PIDNet-L in terms of both performance and speed, although it is slightly slower than SCTNet-B-Seg100. Furthermore, GCNet-L achieves the highest performance with 79.6\% mIoU. As shown in the table, despite GCNet having a higher number of parameters and GFLOPs compared to other models, these factors are not the decisive determinants of inference speed. In fact, inference speed is also influenced by the frequency of memory accesses within the model. The single-path block design of GCNet reduces the number of memory accesses, thereby enhancing inference speed. We also visualized the segmentation results of PIDNet-S, SCTNet-S-Seg75, and GCNet-S, as illustrated in Figure~\ref{vis_seg}. The figure indicates that GCNet exhibits higher accuracy in pixel classification.

\noindent\textbf{CamVid.} The experimental results on the CamVid dataset are shown in Table~\ref{tab_camvid_voc}. To ensure a fair comparison, we retrained the models listed in the table using the MMSegmentation framework and evaluated their inference speeds. The experimental results indicate that our GCNet-S achieves 76.6\% mIoU while also reaching 210.6 FPS. Although PIDNet-S has an mIoU that is 0.6\% higher than that of GCNet-S, its FPS is lower by 67.8. Notably, GCNet-L performs exceptionally well at a resolution of 720 $\times$ 960. Compared to GCNet-S, while its FPS only decreases by 40.2, it achieves 79.1\% mIoU, surpassing all models in the PIDNet series in both performance and speed.

\noindent\textbf{Pascal VOC 2012.} The experimental results on the Pascal VOC 2012 dataset are shown in Table~\ref{tab_camvid_voc}. To ensure a fair comparison, we retrained the models listed in the table using the MMSegmentation framework and evaluated their inference speeds. The experimental results indicate that our GCNet-S achieves 54.9\% mIoU while also reaching 196.9 FPS. Similar to the results on the CamVid dataset, GCNet-L demonstrates remarkable performance and speed at a relatively low resolution (512 $\times$ 2048), even outperforming PIDNet-S. Interestingly, we found that BiSeNetV1 performs better on the VOC dataset, exceeding BiSeNetV2 by 6\% in mIoU, with an FPS that is only 14.3 lower. While BiSeNetV1 excels on VOC, the overall performance across the three datasets shows that GCNet outperforms BiSeNetV1.

\begin{figure}[t]
\centering
\includegraphics[width=\linewidth]{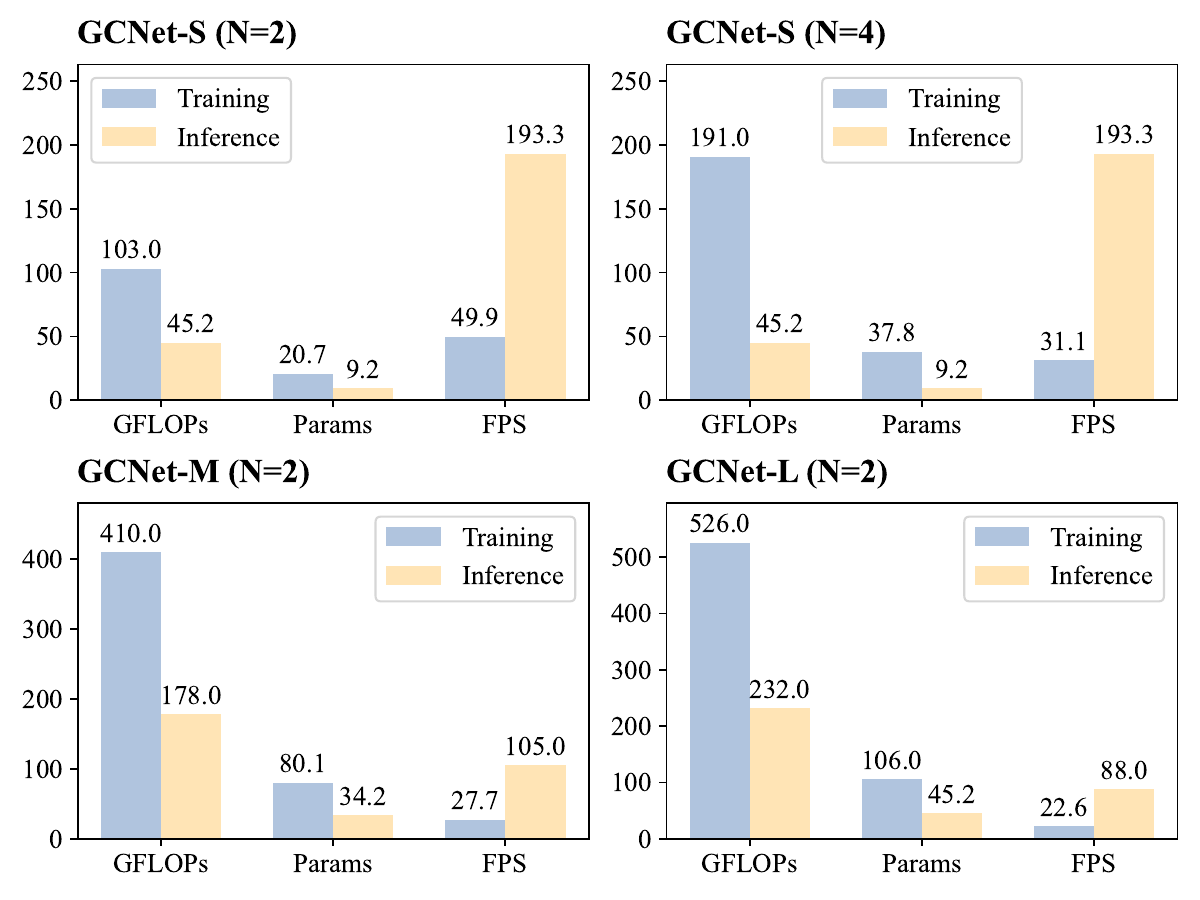}
\caption{The comparison of GFLOPs, number of parameters, and FPS for GCNet during training versus inference.}
\label{training2inference}
\end{figure}

\subsection{Comparison of Training and Inference} 

To visually compare the different configurations of GCNet, we calculated the GFLOPs, number of parameters, and FPS during training versus those during inference, as shown in Figure~\ref{training2inference}. The figure illustrates that GCNet requires significantly more computational resources during training and operates at a slower speed. However, after self-contracting, both GFLOPs and number of parameters are greatly reduced, while FPS experiences a substantial increase. Notably, the performance of both configurations remains equivalent.

\section{Conclusion}
\label{sec:conclusion}

In this study, we proposed GCNet for real-time semantic segmentation. GCNet achieves stronger performance and faster speed simultaneously through self-enlargement and self-contraction, as confirmed by experiments on three public datasets. Although GCNet operates at high speed, its parameter count and computational complexity are higher than those of other models, which limits its applicability on devices with constrained storage capacity. In the future, we plan to explore more powerful and parameter-efficient reparameterizable structures.

\noindent\textbf{Acknowledgment.} This work is supported by Natural Science Foundation China (NSFC) No. 62372301 and No. U22A2097.

{
    \small
    \bibliographystyle{ieeenat_fullname}
    \bibliography{GCNet}

\begin{thebibliography}{38}
\providecommand{\natexlab}[1]{#1}
\providecommand{\url}[1]{\texttt{#1}}
\expandafter\ifx\csname urlstyle\endcsname\relax
  \providecommand{\doi}[1]{doi: #1}\else
  \providecommand{\doi}{doi: \begingroup \urlstyle{rm}\Url}\fi

\bibitem[Brostow et~al.(2009)Brostow, Fauqueur, and Cipolla]{brostow2009semantic}
Gabriel~J Brostow, Julien Fauqueur, and Roberto Cipolla.
\newblock Semantic object classes in video: A high-definition ground truth database.
\newblock \emph{Pattern recognition letters}, 30\penalty0 (2):\penalty0 88--97, 2009.

\bibitem[Cai et~al.(2021)Cai, Dai, Wang, and Li]{cai2021multi}
Yingfeng Cai, Lei Dai, Hai Wang, and Zhixiong Li.
\newblock Multi-target pan-class intrinsic relevance driven model for improving semantic segmentation in autonomous driving.
\newblock \emph{IEEE Transactions on Image Processing}, 30:\penalty0 9069--9084, 2021.

\bibitem[Chen et~al.(2017{\natexlab{a}})Chen, Papandreou, Kokkinos, Murphy, and Yuille]{chen2017deeplab}
Liang-Chieh Chen, George Papandreou, Iasonas Kokkinos, Kevin Murphy, and Alan~L Yuille.
\newblock Deeplab: Semantic image segmentation with deep convolutional nets, atrous convolution, and fully connected crfs.
\newblock \emph{IEEE transactions on pattern analysis and machine intelligence}, 40\penalty0 (4):\penalty0 834--848, 2017{\natexlab{a}}.

\bibitem[Chen et~al.(2017{\natexlab{b}})Chen, Papandreou, Schroff, and Adam]{chen2017rethinking}
Liang-Chieh Chen, George Papandreou, Florian Schroff, and Hartwig Adam.
\newblock Rethinking atrous convolution for semantic image segmentation.
\newblock \emph{arXiv preprint arXiv:1706.05587}, 2017{\natexlab{b}}.

\bibitem[Chen et~al.(2018)Chen, Zhu, Papandreou, Schroff, and Adam]{chen2018encoder}
Liang-Chieh Chen, Yukun Zhu, George Papandreou, Florian Schroff, and Hartwig Adam.
\newblock Encoder-decoder with atrous separable convolution for semantic image segmentation.
\newblock In \emph{Proceedings of the European conference on computer vision (ECCV)}, pages 801--818, 2018.

\bibitem[Cheng et~al.(2021)Cheng, Schwing, and Kirillov]{cheng2021per}
Bowen Cheng, Alex Schwing, and Alexander Kirillov.
\newblock Per-pixel classification is not all you need for semantic segmentation.
\newblock \emph{Advances in neural information processing systems}, 34:\penalty0 17864--17875, 2021.

\bibitem[Cheng et~al.(2022)Cheng, Misra, Schwing, Kirillov, and Girdhar]{cheng2022masked}
Bowen Cheng, Ishan Misra, Alexander~G Schwing, Alexander Kirillov, and Rohit Girdhar.
\newblock Masked-attention mask transformer for universal image segmentation.
\newblock In \emph{Proceedings of the IEEE/CVF conference on computer vision and pattern recognition}, pages 1290--1299, 2022.

\bibitem[Contributors(2020)]{mmseg2020}
MMSegmentation Contributors.
\newblock {MMSegmentation}: Openmmlab semantic segmentation toolbox and benchmark.
\newblock 2020.

\bibitem[Cordts et~al.(2016)Cordts, Omran, Ramos, Rehfeld, Enzweiler, Benenson, Franke, Roth, and Schiele]{cordts2016cityscapes}
Marius Cordts, Mohamed Omran, Sebastian Ramos, Timo Rehfeld, Markus Enzweiler, Rodrigo Benenson, Uwe Franke, Stefan Roth, and Bernt Schiele.
\newblock The cityscapes dataset for semantic urban scene understanding.
\newblock In \emph{Proceedings of the IEEE conference on computer vision and pattern recognition}, pages 3213--3223, 2016.

\bibitem[Deng et~al.(2009)Deng, Dong, Socher, Li, Li, and Fei-Fei]{deng2009imagenet}
Jia Deng, Wei Dong, Richard Socher, Li-Jia Li, Kai Li, and Li Fei-Fei.
\newblock Imagenet: A large-scale hierarchical image database.
\newblock In \emph{2009 IEEE conference on computer vision and pattern recognition}, pages 248--255. Ieee, 2009.

\bibitem[Ding et~al.(2019)Ding, Guo, Ding, and Han]{ding2019acnet}
Xiaohan Ding, Yuchen Guo, Guiguang Ding, and Jungong Han.
\newblock Acnet: Strengthening the kernel skeletons for powerful cnn via asymmetric convolution blocks.
\newblock In \emph{Proceedings of the IEEE/CVF international conference on computer vision}, pages 1911--1920, 2019.

\bibitem[Ding et~al.(2021{\natexlab{a}})Ding, Zhang, Han, and Ding]{ding2021diverse}
Xiaohan Ding, Xiangyu Zhang, Jungong Han, and Guiguang Ding.
\newblock Diverse branch block: Building a convolution as an inception-like unit.
\newblock In \emph{Proceedings of the IEEE/CVF conference on computer vision and pattern recognition}, pages 10886--10895, 2021{\natexlab{a}}.

\bibitem[Ding et~al.(2021{\natexlab{b}})Ding, Zhang, Ma, Han, Ding, and Sun]{ding2021repvgg}
Xiaohan Ding, Xiangyu Zhang, Ningning Ma, Jungong Han, Guiguang Ding, and Jian Sun.
\newblock Repvgg: Making vgg-style convnets great again.
\newblock In \emph{Proceedings of the IEEE/CVF conference on computer vision and pattern recognition}, pages 13733--13742, 2021{\natexlab{b}}.

\bibitem[Everingham et~al.(2010)Everingham, Van~Gool, Williams, Winn, and Zisserman]{everingham2010pascal}
Mark Everingham, Luc Van~Gool, Christopher~KI Williams, John Winn, and Andrew Zisserman.
\newblock The pascal visual object classes (voc) challenge.
\newblock \emph{International journal of computer vision}, 88:\penalty0 303--338, 2010.

\bibitem[Fan et~al.(2021)Fan, Lai, Huang, Wei, Chai, Luo, and Wei]{fan2021rethinking}
Mingyuan Fan, Shenqi Lai, Junshi Huang, Xiaoming Wei, Zhenhua Chai, Junfeng Luo, and Xiaolin Wei.
\newblock Rethinking bisenet for real-time semantic segmentation.
\newblock In \emph{Proceedings of the IEEE/CVF conference on computer vision and pattern recognition}, pages 9716--9725, 2021.

\bibitem[Fu et~al.(2019)Fu, Liu, Tian, Li, Bao, Fang, and Lu]{fu2019dual}
Jun Fu, Jing Liu, Haijie Tian, Yong Li, Yongjun Bao, Zhiwei Fang, and Hanqing Lu.
\newblock Dual attention network for scene segmentation.
\newblock In \emph{Proceedings of the IEEE/CVF conference on computer vision and pattern recognition}, pages 3146--3154, 2019.

\bibitem[He et~al.(2016)He, Zhang, Ren, and Sun]{he2016deep}
Kaiming He, Xiangyu Zhang, Shaoqing Ren, and Jian Sun.
\newblock Deep residual learning for image recognition.
\newblock In \emph{Proceedings of the IEEE conference on computer vision and pattern recognition}, pages 770--778, 2016.

\bibitem[Hu et~al.(2018)Hu, Shen, and Sun]{hu2018squeeze}
Jie Hu, Li Shen, and Gang Sun.
\newblock Squeeze-and-excitation networks.
\newblock In \emph{Proceedings of the IEEE conference on computer vision and pattern recognition}, pages 7132--7141, 2018.

\bibitem[Nirkin et~al.(2021)Nirkin, Wolf, and Hassner]{nirkin2021hyperseg}
Yuval Nirkin, Lior Wolf, and Tal Hassner.
\newblock Hyperseg: Patch-wise hypernetwork for real-time semantic segmentation.
\newblock In \emph{Proceedings of the IEEE/CVF Conference on Computer Vision and Pattern Recognition}, pages 4061--4070, 2021.

\bibitem[Orsic et~al.(2019)Orsic, Kreso, Bevandic, and Segvic]{orsic2019defense}
Marin Orsic, Ivan Kreso, Petra Bevandic, and Sinisa Segvic.
\newblock In defense of pre-trained imagenet architectures for real-time semantic segmentation of road-driving images.
\newblock In \emph{Proceedings of the IEEE/CVF Conference on Computer Vision and Pattern Recognition}, pages 12607--12616, 2019.

\bibitem[Pan et~al.(2022)Pan, Hong, Sun, and Jia]{pan2022deep}
Huihui Pan, Yuanduo Hong, Weichao Sun, and Yisong Jia.
\newblock Deep dual-resolution networks for real-time and accurate semantic segmentation of traffic scenes.
\newblock \emph{IEEE Transactions on Intelligent Transportation Systems}, 24\penalty0 (3):\penalty0 3448--3460, 2022.

\bibitem[Peng et~al.(2022)Peng, Liu, Tang, Hao, Chu, Chen, Wu, Chen, Yu, Du, et~al.]{peng2022pp}
Juncai Peng, Yi Liu, Shiyu Tang, Yuying Hao, Lutao Chu, Guowei Chen, Zewu Wu, Zeyu Chen, Zhiliang Yu, Yuning Du, et~al.
\newblock Pp-liteseg: A superior real-time semantic segmentation model.
\newblock \emph{arXiv preprint arXiv:2204.02681}, 2022.

\bibitem[Poudel et~al.(2019)Poudel, Liwicki, and Cipolla]{poudel2019fast}
Rudra~PK Poudel, Stephan Liwicki, and Roberto Cipolla.
\newblock Fast-scnn: Fast semantic segmentation network.
\newblock \emph{arXiv preprint arXiv:1902.04502}, 2019.

\bibitem[Qi et~al.(2023)Qi, Wu, and Chan]{qi2023mdf}
Wenbo Qi, HC Wu, and SC Chan.
\newblock Mdf-net: A multi-scale dynamic fusion network for breast tumor segmentation of ultrasound images.
\newblock \emph{IEEE Transactions on Image Processing}, 2023.

\bibitem[Romera et~al.(2017)Romera, Alvarez, Bergasa, and Arroyo]{romera2017erfnet}
Eduardo Romera, Jos{\'e}~M Alvarez, Luis~M Bergasa, and Roberto Arroyo.
\newblock Erfnet: Efficient residual factorized convnet for real-time semantic segmentation.
\newblock \emph{IEEE Transactions on Intelligent Transportation Systems}, 19\penalty0 (1):\penalty0 263--272, 2017.

\bibitem[Shelhamer et~al.(2017)Shelhamer, Long, and Darrell]{shelhamer2017fully}
Evan Shelhamer, Jonathan Long, and Trevor Darrell.
\newblock Fully convolutional networks for semantic segmentation.
\newblock \emph{IEEE transactions on pattern analysis and machine intelligence}, 39\penalty0 (4):\penalty0 640--651, 2017.

\bibitem[Shrivastava et~al.(2016)Shrivastava, Gupta, and Girshick]{shrivastava2016training}
Abhinav Shrivastava, Abhinav Gupta, and Ross Girshick.
\newblock Training region-based object detectors with online hard example mining.
\newblock In \emph{Proceedings of the IEEE conference on computer vision and pattern recognition}, pages 761--769, 2016.

\bibitem[Simonyan and Zisserman(2014)]{simonyan2014very}
Karen Simonyan and Andrew Zisserman.
\newblock Very deep convolutional networks for large-scale image recognition.
\newblock \emph{arXiv preprint arXiv:1409.1556}, 2014.

\bibitem[Wang et~al.(2020)Wang, Wu, Zhu, Li, Zuo, and Hu]{wang2020eca}
Qilong Wang, Banggu Wu, Pengfei Zhu, Peihua Li, Wangmeng Zuo, and Qinghua Hu.
\newblock Eca-net: Efficient channel attention for deep convolutional neural networks.
\newblock In \emph{Proceedings of the IEEE/CVF conference on computer vision and pattern recognition}, pages 11534--11542, 2020.

\bibitem[Wu et~al.(2020)Wu, Tang, Zhang, Cao, and Zhang]{wu2020cgnet}
Tianyi Wu, Sheng Tang, Rui Zhang, Juan Cao, and Yongdong Zhang.
\newblock Cgnet: A light-weight context guided network for semantic segmentation.
\newblock \emph{IEEE Transactions on Image Processing}, 30:\penalty0 1169--1179, 2020.

\bibitem[Xie et~al.(2021)Xie, Wang, Yu, Anandkumar, Alvarez, and Luo]{xie2021segformer}
Enze Xie, Wenhai Wang, Zhiding Yu, Anima Anandkumar, Jose~M Alvarez, and Ping Luo.
\newblock Segformer: Simple and efficient design for semantic segmentation with transformers.
\newblock \emph{Advances in neural information processing systems}, 34:\penalty0 12077--12090, 2021.

\bibitem[Xu et~al.(2023)Xu, Xiong, and Bhattacharyya]{xu2023pidnet}
Jiacong Xu, Zixiang Xiong, and Shankar~P Bhattacharyya.
\newblock Pidnet: A real-time semantic segmentation network inspired by pid controllers.
\newblock In \emph{Proceedings of the IEEE/CVF conference on computer vision and pattern recognition}, pages 19529--19539, 2023.

\bibitem[Xu et~al.(2024)Xu, Wu, Yu, Chu, Sang, and Gao]{xu2024sctnet}
Zhengze Xu, Dongyue Wu, Changqian Yu, Xiangxiang Chu, Nong Sang, and Changxin Gao.
\newblock Sctnet: Single-branch cnn with transformer semantic information for real-time segmentation.
\newblock In \emph{Proceedings of the AAAI Conference on Artificial Intelligence}, pages 6378--6386, 2024.

\bibitem[Yu et~al.(2018)Yu, Wang, Peng, Gao, Yu, and Sang]{yu2018bisenet}
Changqian Yu, Jingbo Wang, Chao Peng, Changxin Gao, Gang Yu, and Nong Sang.
\newblock Bisenet: Bilateral segmentation network for real-time semantic segmentation.
\newblock In \emph{Proceedings of the European conference on computer vision (ECCV)}, pages 325--341, 2018.

\bibitem[Yu et~al.(2021)Yu, Gao, Wang, Yu, Shen, and Sang]{yu2021bisenet}
Changqian Yu, Changxin Gao, Jingbo Wang, Gang Yu, Chunhua Shen, and Nong Sang.
\newblock Bisenet v2: Bilateral network with guided aggregation for real-time semantic segmentation.
\newblock \emph{International Journal of Computer Vision}, 129:\penalty0 3051--3068, 2021.

\bibitem[Yu and Koltun(2016)]{YuKoltun2016}
Fisher Yu and Vladlen Koltun.
\newblock Multi-scale context aggregation by dilated convolutions.
\newblock In \emph{ICLR}, 2016.

\bibitem[Zhang et~al.(2021)Zhang, Li, Kosov, Grzegorzek, Shirahama, Jiang, Sun, Li, and Li]{zhang2021lcu}
Jinghua Zhang, Chen Li, Sergey Kosov, Marcin Grzegorzek, Kimiaki Shirahama, Tao Jiang, Changhao Sun, Zihan Li, and Hong Li.
\newblock Lcu-net: A novel low-cost u-net for environmental microorganism image segmentation.
\newblock \emph{Pattern Recognition}, 115:\penalty0 107885, 2021.

\bibitem[Zhao et~al.(2018)Zhao, Qi, Shen, Shi, and Jia]{zhao2018icnet}
Hengshuang Zhao, Xiaojuan Qi, Xiaoyong Shen, Jianping Shi, and Jiaya Jia.
\newblock Icnet for real-time semantic segmentation on high-resolution images.
\newblock In \emph{Proceedings of the European conference on computer vision (ECCV)}, pages 405--420, 2018.

\end{thebibliography}
}

\appendix
\clearpage
\label{sec:appendix}

\section{More Experimental Details} \label{More Experimental Details}

\subsection{Datasets}

\noindent\textbf{Cityscapes.} Cityscapes~\cite{cordts2016cityscapes} is extensively utilized in urban scene understanding and autonomous driving research. It consists of 19 classes and a total of 5000 images, with 2975 designated for training, 500 for validation, and 1525 for testing. The test set is unlabeled, and model predictions must be uploaded to a specific website for evaluation. Each image has been meticulously annotated and possesses a resolution of 1024 $\times$ 2048 pixels.

\noindent\textbf{CamVid.} CamVid~\cite{brostow2009semantic} is the first video dataset to include semantic labels. It comprises 701 images, with 367 designated for training, 101 for validation, and 233 for testing. Each image possesses a resolution of 720 $\times$ 960 pixels and features 32 class labels, though only 11 are typically used for training and evaluation. Notably, many current research~\cite{yu2021bisenet, pan2022deep, xu2023pidnet} utilize the training and validation sets for training and the test set for evaluation, and we adopt this same strategy.

\noindent\textbf{Pascal VOC 2012.} Pascal VOC 2012~\cite{everingham2010pascal} is primarily utilized for image classification, object detection, and image segmentation tasks. It includes 20 classes along with 1 background class. For the semantic segmentation task, the dataset consists of 2913 images, with 1464 allocated to the training set and 1449 to the validation set. Unlike Cityscapes and CamVid, the resolution of these images varies.

\subsection{Implementation Details}

\noindent\textbf{Computing Platform.} The hardware of the computing platform comprises an AMD EPYC 7742 CPU and four NVIDIA A100 GPUs. The software stack includes Ubuntu 20.04.6, CUDA 11.3, PyTorch 1.12.1, TorchVision 0.13.1, MMEngine 0.10.2, and MMSegmentation 1.2.2~\cite{mmseg2020}. During the training phase, both GPUs are utilized, while for the inference phase, only a single GPU is used, with the batch size set to 1.

\noindent\textbf{Training.} Stochastic Gradient Descent (SGD) with a momentum of 0.9 and a weight decay of 0.0005 was employed as the optimizer. Additionally, a polynomial learning rate decay strategy with a power of 0.9 was utilized. For data augmentation, we applied random scaling within the range of 0.5 to 2.0, random cropping, and random flipping with a probability of 0.5. The training iterations, initial learning rate, random crop size, and batch size for the Cityscapes, CamVid, and Pascal VOC 2012 datasets were configured as follows: [120000, 0.01, 1024 $\times$ 1024, 12], [7800, 0.001, 720 $\times$ 960, 12], and [24400, 0.001, 512 $\times$ 512, 16], respectively. Notably, our model was not pretrained on the ImageNet~\cite{deng2009imagenet} dataset, and the Cityscapes model weights were used during training on the CamVid and Pascal VOC 2012 datasets.

\noindent\textbf{Inference.} The inference image sizes for the Cityscapes, CamVid, and Pascal VOC 2012 datasets were configured at 1024 $\times$ 2048, 720 $\times$ 960, and 512 $\times$ 2048, respectively. Since the image resolution in the Pascal VOC dataset is not fixed, images are adjusted to an approximate resolution of 512 $\times$ 2048 to maintain the aspect ratio. The inference speed of the same model varies across different CPUs and software environments, even when using the same GPU. To ensure a fair comparison, we made every effort to reimplement the other models.

\begin{figure}[t]
\centering
\includegraphics[width=\linewidth]{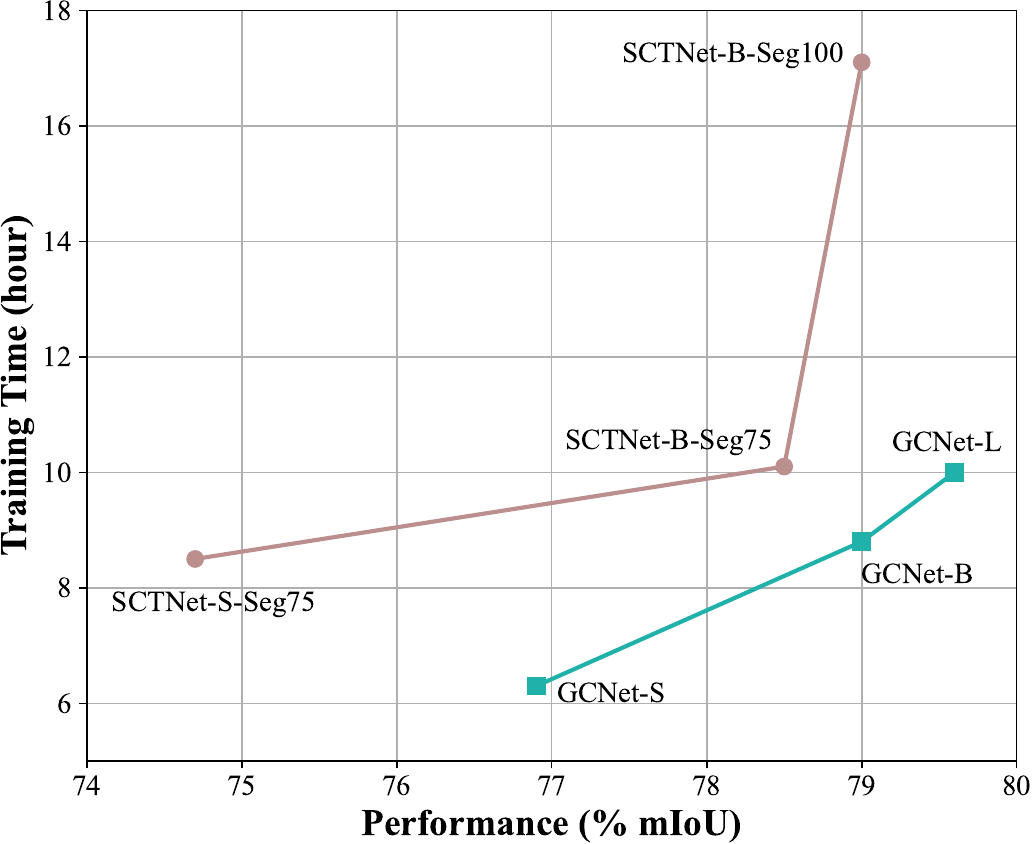}
\caption{Comparison of training time between GCNet and SCTNet on the Cityscapes dataset. Four A100s were used for training and the training time was recorded.}
\label{gcnet_vs_sctnet}
\end{figure}

\begin{figure}[b]
\centering
\includegraphics[width=\linewidth]{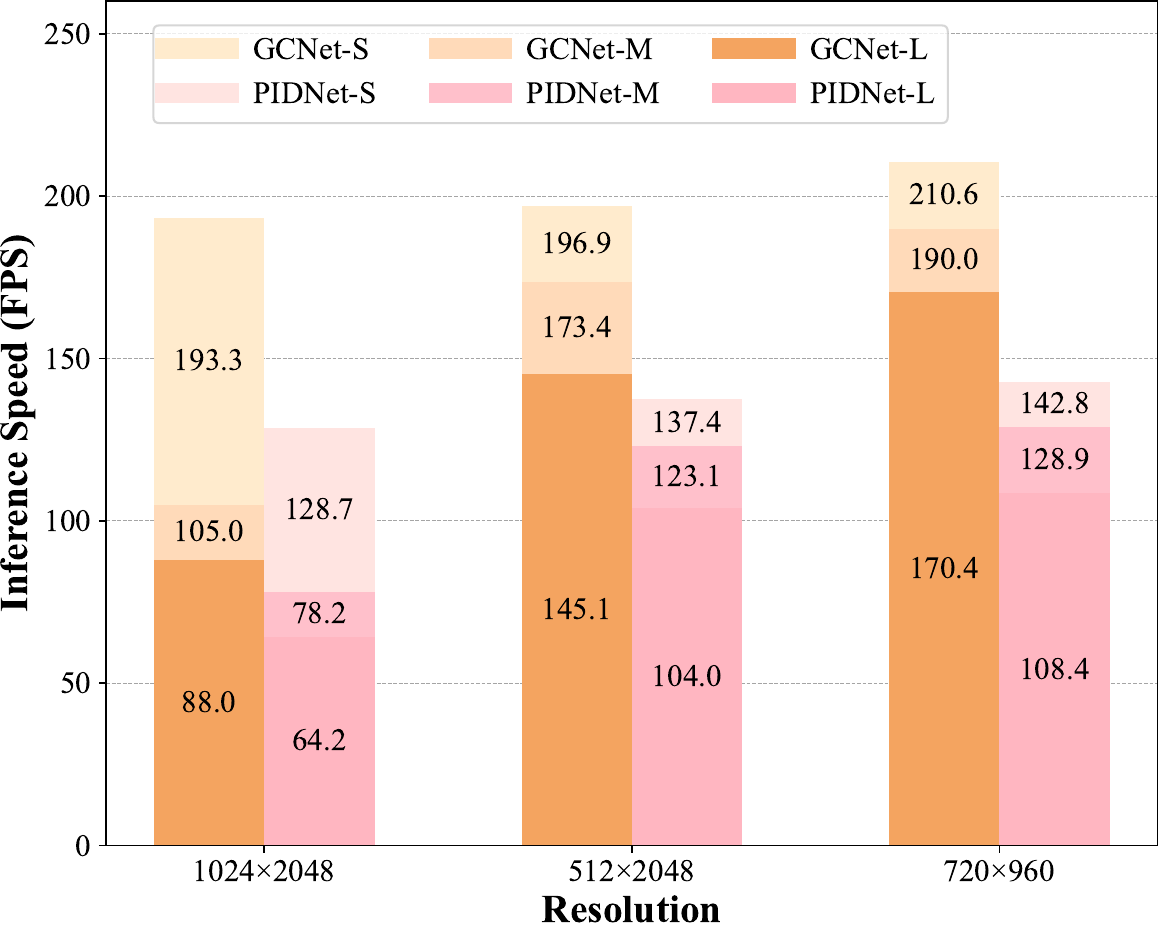}
\caption{Inference speed of different segmentation models for varying resolutions. The GPU used is A100.}
\label{various_resolution}
\end{figure}

\begin{table*}[htbp]
\renewcommand{\arraystretch}{1.12}
\caption{Comparison of inference speed on varying GPUs using the Cityscape validation set. ``*” indicates that the model was reimplemented and retested by us (with convolution and batch normalization fused). ``ImageNet" indicates whether the model utilizes ImageNet pretrained weight.}
\centering
\resizebox{\linewidth}{!}
{
\begin{tabular}{lccccccc} 
\specialrule{0.2em}{0em}{0.3em}
\multirow{2}*{\textbf{Method}}              & \multirow{2}*{\textbf{Resolution}}   & \multirow{2}*{\textbf{RTX 4080}}      & \multirow{2}*{\textbf{RTX 3090}}      & \multirow{2}*{\textbf{V100}}   & \multirow{2}*{\textbf{A100}} & \multirow{2}*{\textbf{ImageNet}}        & \multirow{2}*{\textbf{mIoU}}  \\ 
                                            &                                      &                                       &                                       &                                &                              &                                         &                               \\
\specialrule{0.15em}{0.3em}{0.3em}
ICNet*~\cite{zhao2018icnet}                 & $1024 \times 2048$                   & 92.3                                  & 108.7                                 & 76.7                           & 181.3                        & \ding{55}                               & 71.6                          \\
\specialrule{0.05em}{0.3em}{0.3em}
Fast-SCNN*~\cite{poudel2019fast}            & $1024 \times 2048$                   & 125.3                                 & 211.9                                 & 173.6                          & 325.9                        & \ding{55}                               & 71.0                          \\
\specialrule{0.05em}{0.3em}{0.3em}
CGNet*~\cite{wu2020cgnet}                   & $1024 \times 2048$                   & 38.3                                  & 53.1                                  & 48.5                           & 73.9                         & \ding{55}                               & 68.3                          \\
\specialrule{0.05em}{0.3em}{0.3em}
BiSeNetV1*~\cite{yu2018bisenet}             & $1024 \times 2048$                   & 62.7                                  & 64.3                                  & 56.5                           & 116.8                        & \ding{51}                               & 74.4                          \\
BiSeNetV2*~\cite{yu2021bisenet}             & $1024 \times 2048$                   & 68.7                                  & 69.6                                  & 64.4                           & 132.4                        & \ding{55}                               & 73.6                          \\
\specialrule{0.05em}{0.3em}{0.3em}
STDC1*~\cite{fan2021rethinking}             & $1024 \times 2048$                   & 101.1                                 & 103.3                                 & 89.4                           & 183.7                        & \ding{55}                               & 71.8                          \\
STDC2*~\cite{fan2021rethinking}             & $1024 \times 2048$                   & 75.8                                  & 73.8                                  & 64.7                           & 132.8                        & \ding{55}                               & 74.9                          \\
\specialrule{0.05em}{0.3em}{0.3em}
DDRNet-23-Slim*~\cite{pan2022deep}          & $1024 \times 2048$                   & 94.2                                  & 116.5                                 & 97.6                           & 166.4                        & \ding{55}                               & 76.3                          \\
DDRNet-23*~\cite{pan2022deep}               & $1024 \times 2048$                   & 54.2                                  & 53.4                                  & 47.1                           & 106.0                        & \ding{55}                               & 78.0                          \\
\specialrule{0.05em}{0.3em}{0.3em}
PIDNet-S*~\cite{xu2023pidnet}               & $1024 \times 2048$                   & 60.1                                  & 84.2                                  & 76.3                           & 128.7                        & \ding{55}                               & 76.4                          \\
PIDNet-M*~\cite{xu2023pidnet}               & $1024 \times 2048$                   & 41.7                                  & 41.3                                  & 35.4                           & 78.2                         & \ding{55}                               & 78.2                          \\
PIDNet-L*~\cite{xu2023pidnet}               & $1024 \times 2048$                   & 30.0                                  & 31.1                                  & 27.9                           & 64.2                         & \ding{55}                               & 78.8                          \\
\specialrule{0.05em}{0.3em}{0.3em}
SCTNet-S-Seg50*~\cite{xu2024sctnet}         & $512 \times 1024$                    & 78.9                                  & 124.2                                 & 108.3                          & 169.1                        & \ding{55}                               & 71.0                          \\
SCTNet-S-Seg75*~\cite{xu2024sctnet}         & $768 \times 1536$                    & 78.7                                  & 124.0                                 & 99.7                           & 168.7                        & \ding{55}                               & 74.7                          \\
SCTNet-B-Seg50*~\cite{xu2024sctnet}         & $512 \times 1024$                    & 77.3                                  & 119.0                                 & 97.6                           & 162.6                        & \ding{55}                               & 75.0                          \\
SCTNet-B-Seg75*~\cite{xu2024sctnet}         & $768 \times 1536$                    & 73.9                                  & 101.9                                 & 83.1                           & 157.3                        & \ding{55}                               & 78.5                          \\
SCTNet-B-Seg100*~\cite{xu2024sctnet}        & $1024 \times 2048$                   & 63.6                                  & 63.9                                  & 53.3                           & 117.0                        & \ding{55}                               & 79.0                          \\
\specialrule{0.05em}{0.3em}{0.3em}
\rowcolor{green!25}
GCNet-S                                     & $1024 \times 2048$                   & 110.1                                 & 130.9                                 & 114.1                          & 193.3                        & \ding{55}                               & 77.3                          \\
\rowcolor{green!25}
GCNet-M                                     & $1024 \times 2048$                   & 47.4                                  & 50.1                                  & 44.2                           & 105.0                        & \ding{55}                               & 79.0                          \\
\rowcolor{green!25}
GCNet-L                                     & $1024 \times 2048$                   & 38.3                                  & 40.7                                  & 36.2                           & 88.0                         & \ding{55}                               & 79.6                          \\
\specialrule{0.2em}{0.3em}{0em}
\end{tabular}
}
\label{tab_varying_gpu}
\end{table*}

\subsection{Metrics}

We adopt mIoU (mean Intersection over Union), number of parameters, GFLOPs (Giga Floating Point Operations), and FPS (Frames Per Second) as metrics. mIoU is commonly used in image segmentation tasks, where it measures the average overlap between predicted results and ground truth annotations, serving as an indicator of model performance. Number of parameters refers to the total number of trainable parameters in the model, providing a measure of its size. GFLOPs quantifies the computational load of the model, reflecting its computational complexity. FPS represents the number of image frames the model processes per second, serving as a metric for inference speed. In general, number of parameters and GFLOPs are not the decisive factors influencing FPS. This paper focuses on optimizing mIoU and FPS, rather than number of parameters and GFLOPs.

\section{More Experiments}

\subsection{Training Time of GCNet and SCTNet}

The paper mentions that SCTNet requires a high-performance segmentation model for knowledge distillation training, which is quite time-consuming. To verify that GCNet demands less training time compared to SCTNet, we recorded the training times of various versions of both models, as shown in Figure~\ref{gcnet_vs_sctnet}. Since SCTNet-S/B-Seg50 and SCTNet-S/B-Seg75 share the same training configuration, differing only in inference settings, we only recorded the training time for SCTNet-S-Seg75 and SCTNet-B-Seg75. The figure reveals that GCNet not only requires less training time but also achieves higher performance than SCTNet. Specifically, SCTNet-B-Seg100 takes 17.1 hours to reach 79.0\% mIoU, while GCNet-B achieves this in only 8.8 hours.

\subsection{Inference Speed With Varying Resolution}

To provide an intuitive understanding of the inference speed of GCNet at varying resolutions, we visualized its FPS, as shown in Figure~\ref{various_resolution}. The figure reveals that as resolution decreases, the FPS of both GCNet and PIDNet increases significantly, particularly for the M and L versions. Surprisingly, GCNet-M and GCNet-L achieve outstanding FPS at lower resolutions (512 $\times$ 2048 and 720 $\times$ 960), with GCNet-L even surpassing PIDNet-S. This may be attributed to the benefits of reduced memory access enabled by the single-path block, as memory access represents a significant computational cost in computer systems.

\subsection{Inference Speed With Varying GPUs}

To demonstrate GCNet's versatility across varying GPUs, we conducted speed tests on the RTX 4080, RTX 3090, V100, and A100, as shown in Table~\ref{tab_varying_gpu}. The results reveal that GCNet performs well on both consumer-grade GPUs (RTX 4080 and RTX 3090) and professional-grade GPUs (V100 and A100). Interestingly, smaller models show faster inference speeds on the RTX 3090 compared to the RTX 4080, while larger models perform similarly on both. As a single-branch architecture model, SCTNet is relatively insensitive to changes in lower resolutions, with comparable inference speeds for Seg50 and Seg75. In contrast, multi-branch models show substantial speed improvements with lower resolutions. As illustrated in Figure~\ref{various_resolution}, inference speeds for multi-branch models, especially GCNet, increase significantly as the resolution decreases. We attribute this to the high-resolution branch in multi-branch architectures, which requires the maintenance of larger feature maps and thus more computations. In future work, we plan to further investigate GCNet on lower resolutions using the Cityscapes dataset.

\end{document}